\documentclass[10pt,journal,compsoc]{IEEEtran}

\ifCLASSOPTIONcompsoc
  \usepackage[nocompress]{cite}
\else
  \usepackage{cite}
\fi

\ifCLASSINFOpdf
\else
\fi

\usepackage{times}
\usepackage{epsfig}
\usepackage{graphicx}
\usepackage{amsmath}
\usepackage{amssymb}
\usepackage{footnote}
\usepackage{dsfont}
\usepackage{enumitem}

\usepackage{comment}
\usepackage{blindtext}
\usepackage{morewrites}
\usepackage{makecell}
\usepackage{pifont}%
\usepackage{multirow}
\usepackage{xcolor}
\usepackage{float}

\usepackage[utf8]{inputenc} %
\usepackage[T1]{fontenc}    %
\usepackage{hyperref}       %
\usepackage{url}            %
\usepackage{booktabs}       %
\usepackage{amsfonts}       %
\usepackage{nicefrac}       %
\usepackage{microtype}      %
\usepackage{xcolor}         %

\usepackage[utf8]{inputenc} %
\usepackage[T1]{fontenc}    %
\usepackage{hyperref}       %
\usepackage{url}            %
\usepackage{booktabs}       %
\usepackage{amsfonts}       %
\usepackage{nicefrac}       %
\usepackage{microtype}      %
\usepackage{xcolor}         %

\usepackage{wrapfig}
\usepackage{graphicx}
\usepackage{amsmath}
\usepackage{mathtools}
\usepackage{multirow}
\usepackage{makecell}
\usepackage{tabularx}
\usepackage{algorithm2e}
\usepackage{fancyvrb,xcolor}
\usepackage{nicematrix}
\usepackage{bbm}

\usepackage{algorithm2e}
\usepackage{algorithmic}
\usepackage{fancyvrb}
\usepackage{xcolor}
\usepackage{nicematrix}

\usepackage{tikz}
\usepackage{comment}
\usepackage{amsmath,amssymb} %
\usepackage{color}
\usepackage{enumitem}
\usepackage{amsthm}

\newtheorem{remark}{Remark}
\usepackage{multirow}
\usepackage{makecell}
\usepackage{amsmath}
\usepackage{capt-of}
\usepackage{tabularx}
\usepackage{epsfig}
\usepackage{amssymb}
\usepackage{amsfonts}
\usepackage{booktabs}
\usepackage{scalerel}
\usepackage{listings}
\usepackage{varwidth}
\usepackage[export]{adjustbox}
\usepackage{tikz}
\usetikzlibrary{tikzmark}

\usepackage{stmaryrd}
\usepackage{bbm}
\usepackage{wrapfig}
\usepackage{pifont}
\usepackage[utf8]{inputenc}

\definecolor{deepblue}{rgb}{0,0,0.5}
\definecolor{officeblue}{RGB}{0,102,204}
\definecolor{deepred}{rgb}{0.6,0,0}
\definecolor{deepgreen}{rgb}{0,0.5,0}
\definecolor{mybrickred}{RGB}{182,50,28}

\definecolor{fillcolor}{RGB}{216,217,252}

\newcommand*\AlgCommentInLine[1]{{\color{deepblue}{$\triangleright$ \textit{#1}}}}

\newcommand{\rebuttaltpami}[1]{\textcolor{black}{#1}}

\begin{document}

\title{BRACTIVE: A Brain Activation Approach to Human Visual Brain Learning}

\author{
Xuan-Bac~Nguyen~\IEEEmembership{Student~Member,~IEEE},
Hojin~Jang~\IEEEmembership{Member,~IEEE}, 
Xin~Li~\IEEEmembership{Fellow,~IEEE},
Samee~U.~Khan~\IEEEmembership{Senior Member,~IEEE},
Pawan~Sinha~\IEEEmembership{Member,~IEEE},
Khoa~Luu~\IEEEmembership{Senior Member,~IEEE}
\IEEEcompsocitemizethanks{
\IEEEcompsocthanksitem Xuan-Bac Nguyen and Khoa Luu are with the Electrical Engineering \& Computer Science Department, University of Arkansas, Fayetteville, Arkansas 72703 USA. E-mail: xnguyen@uark.edu, khoaluu@uark.edu. 
\IEEEcompsocthanksitem Hojin Jang is with the Brain and Cognitive Engineering, Korea University, Seoul, South Korea. E-mail: hojin4671@korea.ac.kr.
\IEEEcompsocthanksitem Xin Li is with the Department of Computer Science, University at Albany, Albany, NY 12222 USA. E-mail: xli48@albany.edu.
\IEEEcompsocthanksitem Samee U. Khan is with the Mike Wiegers Department of Electrical \& Computer Engineering, Kansas State University, Manhattan, KS 66506 USA. E-mail: sameekhan@ksu.edu.
\IEEEcompsocthanksitem Pawan Sinha is with the Department of Brain and Cognitive Sciences, Massachusetts Institute of Technology, Cambridge, MA 02139-4307 USA. E-mail: psinha@mit.edu.
}
}

\IEEEtitleabstractindextext{%
\begin{abstract}
The human brain is a highly efficient processing unit, and understanding how it works can inspire new algorithms and architectures in machine learning. In this work, we introduce a novel framework named Brain Activation Network (BRACTIVE), a transformer-based approach to studying the human visual brain. The primary objective of BRACTIVE is to align the visual features of subjects with their corresponding brain representations using functional Magnetic Resonance Imaging (fMRI) signals. It enables us to identify the brain's Regions of Interest (ROIs) in the subjects. Unlike previous brain research methods, which can only identify ROIs for one subject at a time and are limited by the number of subjects, BRACTIVE automatically extends this identification to multiple subjects and ROIs. Our experiments demonstrate that BRACTIVE effectively identifies person-specific regions of interest, such as face and body-selective areas, aligning with neuroscience findings and indicating potential applicability to various object categories. More importantly, we found that leveraging human visual brain activity to guide deep neural networks enhances performance across various benchmarks. It encourages the potential of BRACTIVE in both neuroscience and machine intelligence studies.
\end{abstract}

}

\maketitle

\IEEEdisplaynontitleabstractindextext

\IEEEpeerreviewmaketitle

\section{Introduction}

Artificial intelligence (AI) algorithms \cite{bert, liu2019roberta, dosovitskiy2020image, clip} have recently advanced significantly, bringing capabilities one step closer to a human-like level, thanks to the development of deep learning. Inspired by human brain mechanisms, the feedforward and deep neural networks have greatly enhanced performance across various domains such as computer vision, natural language processing, speech recognition, etc \cite{nguyen2021clusformer, nguyen2020self, nguyen2023micron, nguyen2022multi,truong2022otadapt,nguyen2019sketch,nguyen2022two,nguyen2023fairness,nguyen2023brainformer,nguyen2023insect,nguyen2019audio,serna2024video}. Recent studies show that these methods are more likely to mirror the mechanism of the human brain. For instance, a relationship has been found between brain representations in the visual pathway and the hierarchical layers within Deep Neural Networks (DNNs) \cite{pmid28530228,pmid27282108,gucclu2015deep,yamins2014performance,cichy2016comparison}. They have initiated a research direction involving both cognitive neuroscientists and AI researchers to uncover the mechanisms and complexities of the human brain, with the potential to adapt these findings to AI models \cite{Spampinato2016deep, 8237631, pmid21945275, pmid23932491}.

In one direction, brain-inspired representations offer a promising avenue for enhancing DNN training. For example, incorporating brain data, such as Electroencephalogram (EEG) signals, has been shown to improve performance in classification and salient detection tasks \cite{palazzo2020decoding}. Another study showed that representations simulating the primate primary visual cortex in DNN models significantly enhance robustness to adversarial attacks and image corruptions \cite{dapello2020simulating}. This approach leverages the inherent strengths of neural processing to guide and refine artificial models. Conversely, AI models possess unique capabilities that can significantly advance our understanding of neuroscience \cite{bashivan2019neural, ponce2019evolving, bao2020map}. These models can simulate complex neural interactions, offering insights that are challenging to obtain through traditional experimental methods. Specifically, our research highlights the advantages of integrating deep learning techniques with neuroscience to elucidate the topological representations of objects within the brain. By mapping these representations, we can gain a deeper understanding of how the brain organizes and processes visual information.

\begin{figure*}
    \centering
    \includegraphics[width=1.0\linewidth]{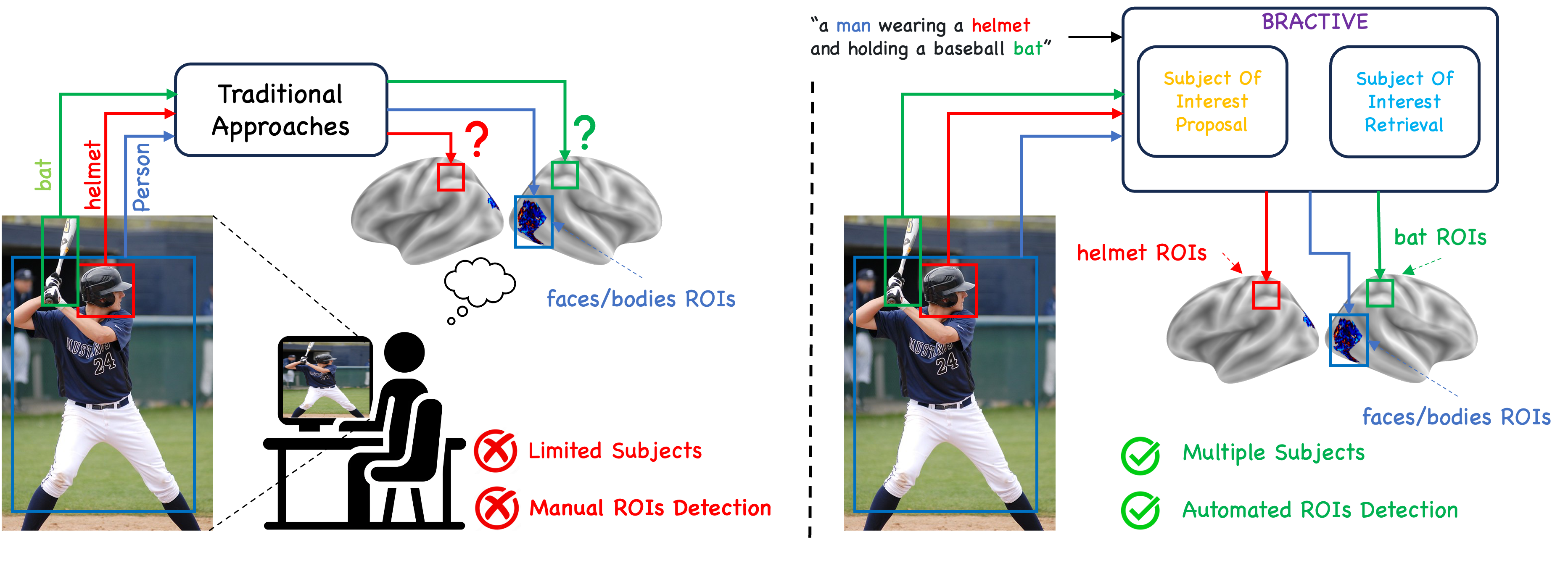}
    \vspace{-10mm}
    \caption{The proposed method can automatically determine the subject of interest and corresponding ROIs inside the brain that the participant is focusing on. }
    \label{fig:abstract_figure}
    \vspace{-4mm}
\end{figure*}

\textbf{Contributions of This Work.}
In summary, the contributions in this paper are four-folds. First, we present a novel transformer-based Brain Activation Network (BRACTIVE)\footnote{The code of this work will be released.} for understanding the human visual brain. Apart from the prior work, our approach can automatically identify the brain's Regions of Interest (ROIs) for multiple subjects, as shown in Fig. \ref{fig:abstract_figure}. Second, we propose a novel module, named Subject of Interest Proposal (SOIP), to predict the subjects that participants focus on during visual tasks. Third, since the labels for the proposed tasks are not available, we introduce the Subject of Interest Retrieval (SOIR) module to establish features of these subjects in visual and fMRI modalities. Finally, the Weighted SOI Loss function is presented to align the features of subjects across modalities.
The experiments show ROIs for humans identified by BRACTIVE are closely aligned with the findings (floc-faces, floc-bodies) from neuroscience. It promises to bring valuable references for neuroscientists in studying visual brain functions.

\section{Related Work}

Decoding human brain representation has been one of the most popular research topics for a decade. In particular, cognitive neuroscience has made substantial advances in understanding neural representations originating in the primary Visual Cortex (V1) \cite{seymour2016representation}. Indeed, the primary visual context responds to the processing of information related to oriented edges and colors. The V1 forwards the information to other neural regions, focusing more on complex shapes and features. These regions are overlapped mainly with receptive fields such as V4 \cite{peirce2015understanding}, before converging on object and category representations in the inferior temporal (IT) cortex \cite{cortex2005fast}. Neuroimaging techniques, including Functional Magnetic Resonance Imaging (fMRI), Magnetoencephalography (MEG), and electroencephalogram (EEG), have been crucial in these studies. However, to replicate human-level neural representations that fully capture our visual processes, it is crucial to precisely monitor the activity of every neuron in the brain simultaneously. Consequently, recent efforts in brain representation decoding have focused on exploring the correlation between neural activity data and computational models. In this research direction, several studies \cite{haynes2005predicting, thirion2006inverse, kamitani2005decoding, cox2003functional, haxby2001distributed} were presented to decode brain information. Recently, with the help of deep learning, authors in \cite{chen2023seeing, scotti2023reconstructing, takagi2023high, ozcelik2023natural, lin2022mind, ozcelik2023brain} have presented methods to reconstruct what humans see from fMRI signals using diffusion models. The authors in \cite{kim2023swift, chen2023seeing} also explored patterns of fMRI signals. However, they often fall short of demonstrating or explaining the nature of these patterns.

In addition, there are recent studies on the intersection between AI and cognitive neuroscience \cite{thomas2022self, thual2022aligning, kan2022brain, portes2022distinguishing, lin2022mind, millet2022toward, fong2018using, safarani2021towards, 10073607, pogoncheff2023explaining, figclip, buch2022revisiting, fang2021clip2video, luo2021clip4clip, ma2022x, qing2023disentangling, rasheed2023fine, wang2021actionclip, wang2023unified, xu2021videoclip, xue2023clip, khosla2022characterizing, cui2022fuzzy, sarch2023brain, duan2024few, lee2024hyper, thomas2022self, quesada2022mtneuro,nguyen2023brainformer,nguyen2024diffusion,nguyen2024hierarchical,nguyen2024qclusformer,nguyen2024quantum,nguyen2024quantumbrain}. Deep learning methods \cite{nsd, 27yamins2013hierarchical,28yamins2014performance,29kriegeskorte2008representational, nguyen2023algonauts} have been applied to predict the neural responses. Other studies, inspired by biological mechanisms such as memory \cite{30graves2016hybrid}, coding theory, and attention \cite{31gregor2015draw,32xu2015show}, are increasingly being adopted in the AI field. Some recent methods have used neural activity data to guide the training of models \cite{palazzo2020decoding, 34fong2018using,Spampinato2016deep,nishida2020brain,nechyba1995human}. They utilized EEG and fMRI signals to constrain the neural network to behave in a manner similar to the neural response in the visual cortex.

\textbf{Discussion:} Neuroscience and computer science mutually benefit each other. Insights from the human cognitive system can lead to innovative AI algorithms, while advanced deep neural networks can enhance our exploration of the human brain. 
By delving deeper into the intricacies of how the brain processes information, learns, and makes decisions, researchers are paving the way for a new era of AI. This understanding can be translated into more efficient AI solutions that require less training data and can learn and adapt in ways that mirror human learning. 
According to neuroscience findings, V1 is responsible for edge and orientation processing, while V4 can handle larger receptive fields. The object-specific ROIs such as \textit{floc-faces} and \textit{floc-bodies} are responsible for recognizing human subjects. \textit{This prompts the question: Which Regions of Interest (ROIs) are related to processing information of other subjects, such as animals, bicycles, cars, etc?}. Therefore, this paper focuses on understanding how the human visual brain processes visual stimuli. Apart from the prior studies, our approach is not limited by predefined visual ROI from the neuroscience field. Instead, the proposed method can automatically explore more ROI for a specific subject.

\section{The Proposed Brain Activation Network Approach}

\begin{figure*}
    \centering
    \includegraphics[width=1.0\linewidth]{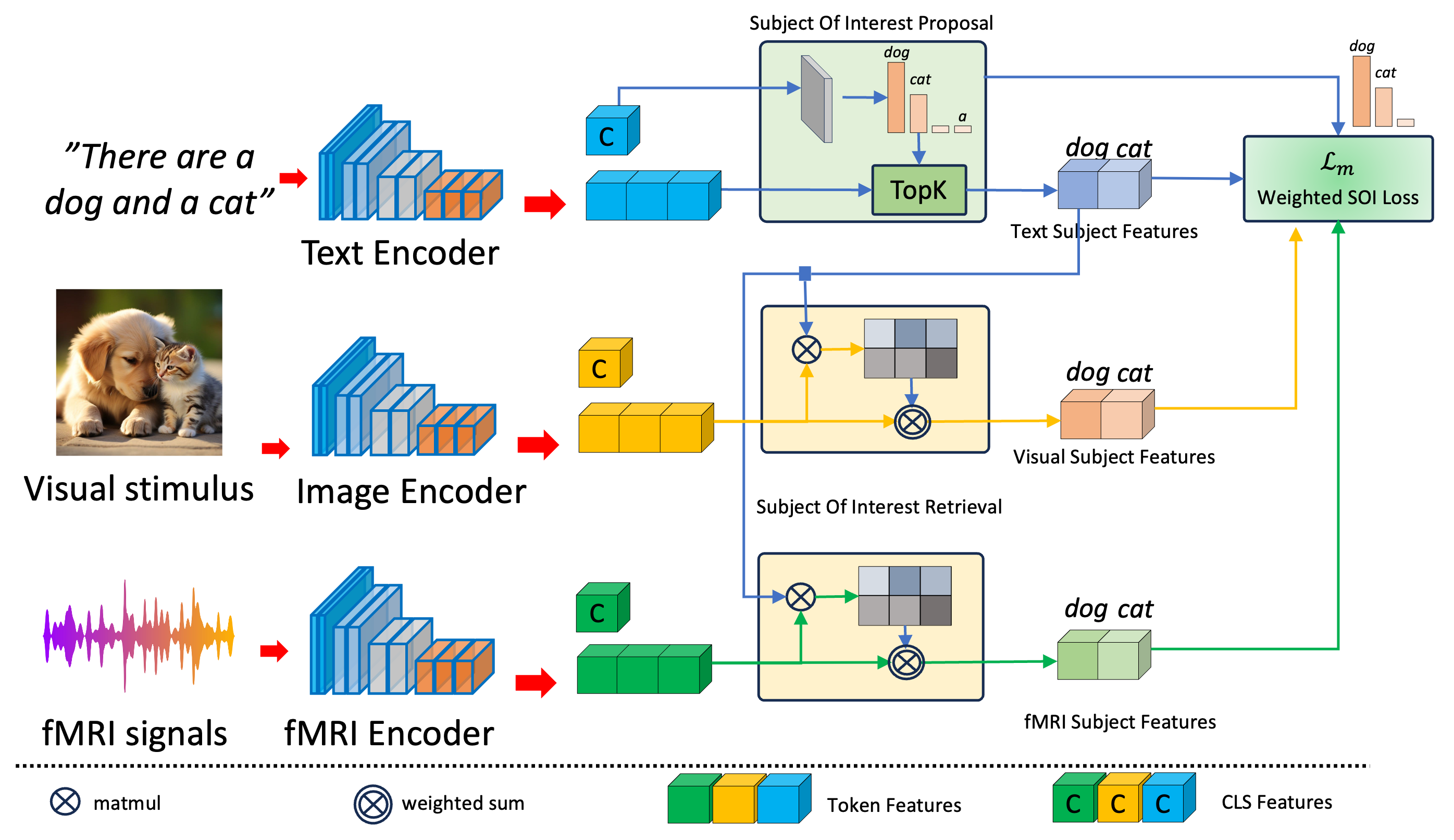}
    \vspace{-10mm}
    \caption{The illustration of BRACTIVE and training objectives. The Subject of Interest Proposal identifies the interested subjects that participants are focusing on. The Subject of Interest Retrieval is used to construct the corresponding features of the proposed subjects on visual and fMRI modalities. The Weighted SOI Loss is used to align the features of these subjects.}
    \label{fig:overview_bractive}
    \vspace{-4mm}
\end{figure*}

In this section, we detail the proposed method as illustrated in Fig. \ref{fig:overview_bractive}. In particular, we first briefly introduce the vision encoder, fMRI encoder, and text encoder in Subsections \ref{subsec:vision_encoder}, \ref{subsec:fmri_encoder}, \ref{subsec:text_encoder}, respectively. Then, the details of the Subject of Interest Proposal, Subject of Interest Retrieval modules, and overall objective loss functions are presented in Subsections \ref{subsec:soip}, \ref{subsec:soir}, \ref{subsec:objective_loss_functions}, correspondingly.

\subsection{Vision Encoder}
\label{subsec:vision_encoder}
Given a visual stimulus $x \in \mathbb{R}^{h \times w \times c}$, we split $x$ into multiple non-overlap patches $\mathcal{P} = \{p_i\}_{i=0}^{N_p -1}$ where $N_p = |\mathcal{P}| = (h \times w) / {p^2}$ is the number of patches, and $p$ is the patch size. 
Let $\texttt{Enc}_v$ be a visual encoder, i.e., ViT, which receives $\mathcal{P}$ as the input. The vectors of latent and patches are represented as in Eq. \eqref{eq:visual_encoder}.
\begin{equation}
    \label{eq:visual_encoder}
    \begin{split}
    \mathbf{P}_{\texttt{CLS}}, \left[\mathbf{P}_0, \dots \mathbf{P}_i, \dots \mathbf{P}_{N_p - 1}\right] = \texttt{Enc}_v(\mathcal{P}) \\ \mathbf{P}_{\texttt{CLS}} \in \mathbb{R}^{1 \times d}, \mathbf{P}_{i} \in \mathbb{R}^{1 \times d}
    \end{split}
\end{equation}
where $\mathbf{P}_{i}$ is the feature vector of the $i^{th}$ patch and $\mathbf{P}_{\texttt{CLS}}$ is the feature vector representing context in the visual stimulus $x$. Since $\texttt{Enc}_v$ is a transformer-based architecture, we can write $\mathbf{P}_{\texttt{CLS}} = \sum_{i=0}^{N_p-1} \alpha_i \mathbf{P}_i$ where $\sum \alpha_i = 1$ and $0 \leq \alpha_i \leq 1$. Similarly, consider a subject $s_k$ appears in $x$, the visual feature of $s_k$ can be represented in the Eqn \eqref{eq:visual_encoder} %
\begin{equation}
    \label{eq:visual_object}
    \mathbf{P}_{s_k} = \sum_{i=0}^{N_p-1} \alpha_{ik} \mathbf{P}_i
\end{equation}
where $\alpha_{ik}$ acts as a coefficient, expressing how much the texture of $s_k$ is included in $p_i$.

\begin{remark}
\label{rm1}
By finding the coefficient $\alpha_{ik}$, we can determine the regions or patches in the image that correspond to $\mathbf{P}_{s_k}$.
\end{remark}

\subsection{fMRI Encoder}
\label{subsec:fmri_encoder}
In dealing with the high-dimensional fMRI 1D signal $f \in \mathbb{R}^{N_F}$, where $N_F$ might reach 100K, a direct strategy involves employing linear or fully connected layers \cite{scotti2023reconstructing} to model the signal and derive its latent embedding. However, this approach encounters two main limitations. First, fully connected layers struggle with high-dimensional features, making it difficult to extract meaningful information while burdening the model's memory usage. Second, the current representation form of the fMRI, as a signal, makes it difficult to explore the local patterns of nearby voxels \rebuttaltpami{(Please refer to the Appendix/Supplementary for the details)}. To address these challenges, we project these signals into a flattened 2D space by leveraging the transformation function of \textit{fsaverage} surface. Then we have $f_{2d} = \texttt{T}_{2d}(f) \in \mathbb{R}^{h_{2d} \times w_{2d}}$
where $f_{2d}$ is the flattened 2D form of the fMRI signal $f$, $h_{2d}$ and $w_{2d}$ are the height and width, correspondingly. Consequently, we apply a transformer-based encoder, denoted $\texttt{Enc}_f$, to extract its feature. The output features of the fMRI signals are represented as in Eq. \eqref{eq:fmri_encoder}.
\begin{equation}
    \label{eq:fmri_encoder}
    \begin{split}
            \mathbf{F}_{\texttt{CLS}}&, \left[\mathbf{F}_0, \dots \mathbf{F}_i, \dots \mathbf{F}_{N_r - 1}\right] = \texttt{Enc}_f(f_{2d}) \\
            \mathbf{F}_{\texttt{CLS}}& \in \mathbb{R}^{1 \times d}, \mathbf{F}_{i} \in \mathbb{R}^{1 \times d}
    \end{split}
\end{equation}
Similar to the visual encoder, 
the \textit{fMRI features} of the subject $s_k$ is defined as in Eqn \eqref{eq:fmri_object} %

\begin{equation}
    \label{eq:fmri_object}
    \mathbf{F}_{s_k} = \sum_{i=0}^{N_r-1} \beta_{ik} \mathbf{F}_i
\end{equation}

\begin{remark}
\label{rm2}
By finding the coefficient $\beta_{ik}$, we can determine the sub-signals/voxels in the fMRI that correspond to $\mathbf{F}_{s_k}$.
\end{remark}

\begin{figure*}[!t]
    \centering
    \includegraphics[width=0.95\linewidth]{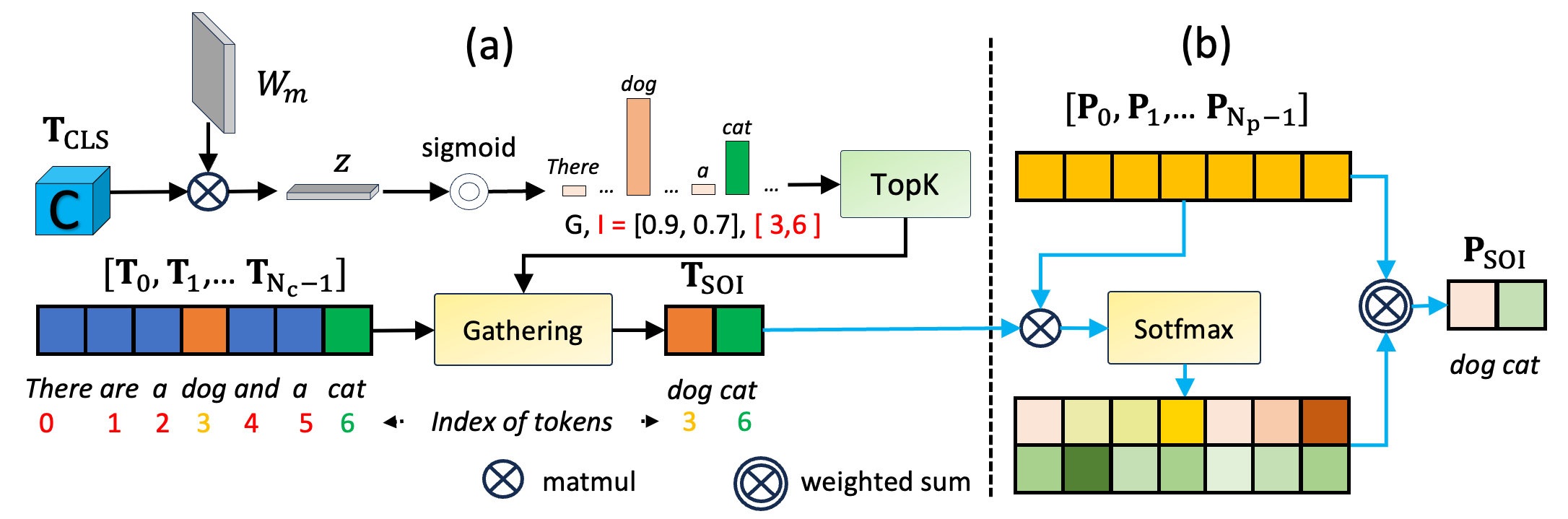}
    \vspace{-4mm}
    \caption{Detailed illustration of (a) Subject of Interest Proposal (SOIP) and (b)  Subject of Interest Retrieval (SOIR) module. It is similar to applying the SOIR module for the fMRI modality by replacing vision tokens $\textbf{P}_i$ by $\textbf{F}_i$, respectively.}
    \label{fig:msp}
    \vspace{-4mm}
\end{figure*}

\subsection{Text Encoder}
\label{subsec:text_encoder}
Since $\mathbf{P}_{s_k}$ and $\mathbf{F}_{s_k}$ represent the visual and fMRI modalities of the $s_k$ subject, respectively, we need semantic labels in visual to construct these features. However, we require prior access to this information, which poses a significant obstacle. 
To address this challenge, we propose utilizing the text modality to represent the $s_k$ for the following reasons. First, unlike the visual and fMRI modalities that require \textbf{multiple} patches or pixels and sub-signals/voxels to represent the subject, a \textbf{single} word is sufficient to describe the subject (e.g., cat, dog, etc.) in the text modality. Secondly, the features of the text are more discriminative than those of images, thanks to the BERT-based architecture \cite{bert}. Thus, text poses a strong representation of a subject. 

Let $t$ be the text description describing the visual stimulus context $x$. We assume that $t$ contains the word represent for the subjects inside $x$. Following prior work \cite{clip,dosovitskiy2020image}, we use transformer-based architecture to extract the features of $t$ as following:
\begin{equation}
    \label{eq:text_encoder}
    \begin{split}
           \mathbf{T}_{\texttt{CLS}}&, \left[\mathbf{T}_0, \dots \mathbf{T}_i, \dots \mathbf{T}_{N_c - 1}\right] = \texttt{Enc}_t(\mathbf{t}) \\ \mathbf{T}_{\texttt{CLS}}& \in \mathbb{R}^{1 \times d}, \mathbf{T}_{i} \in \mathbb{R}^{1 \times d} 
    \end{split}
\end{equation}
Similar to visual and fMRI modalities, we have
$\textbf{T}_{s_k}$ to represent context feature and \textit{text features} of $s_k$, respectively. However, $\textbf{T}_{s_k}$ is not a combination of any tokens in the $t$, instead, $\textbf{T}_{s_k}\in \left[\mathbf{T}_0, \dots \mathbf{T}_i, \dots \mathbf{T}_{N_c - 1}\right]$ is one of tokens in $t$. In the next section, we present the new \textit{Subject of Interest Proposal} module to determine which token represents the $s_k$.

\subsection{Subject of Interest Proposal}
\label{subsec:soip}
\begin{algorithm}[!ht]
\centering
\small
\caption{Pseudo code of BRACTIVE.}
\label{algo:BRACTIVE}
\begin{algorithmic}[1]
\STATE {\bfseries Input:} The visual stimulus $x$, fMRI signal $f$, \\ text description $t$. Let $k$ be the number of the subject \\of interest proposal. 
\STATE {\bfseries Output:} The objective loss.

\AlgCommentInLine{Extract visual features}
\STATE{$\mathcal{P} \gets \texttt{Patchify}(x)$} 
\STATE{$\mathbf{P}_{\texttt{CLS}}, \left[\mathbf{P}_0, \dots \mathbf{P}_i, \dots \mathbf{P}_{N_p - 1}\right] \gets \texttt{Enc}_v(\mathcal{P})$}

\AlgCommentInLine{Extract fMRI features}
\STATE{$f_{2d} \gets \texttt{T}_{2d}(f)$}
\STATE{$\mathbf{F}_{\texttt{CLS}}, \left[\mathbf{F}_0, \dots \mathbf{F}_i, \dots \mathbf{F}_{N_r - 1}\right] \gets \texttt{Enc}_f(f_{2d})$}

\AlgCommentInLine{Extract text features}
\STATE{$\mathbf{T}_{\texttt{CLS}}, \left[\mathbf{T}_0, \dots \mathbf{T}_i, \dots \mathbf{T}_{N_c - 1}\right] \gets \texttt{Enc}_t(t)$}

\AlgCommentInLine{Subject of interest proposal}
\STATE{$\mathbf{T}_{\texttt{SOI}}, G \gets \texttt{SOIP}(\mathbf{T}_{\texttt{CLS}}, k)$}

\AlgCommentInLine{Subject of interest retrieval}
\STATE{$\mathbf{P}_{\texttt{SOI}} \gets \texttt{SOIR}(\mathbf{T}_{\texttt{SOI}}, \mathbf{P}_v)$}
\STATE{$\mathbf{T}_{\texttt{SOI}} \gets \texttt{SOIR}(\mathbf{F}_{\texttt{SOI}}, \mathbf{F}_f)$}

\AlgCommentInLine{Global contextual loss}
\STATE{$\mathcal{L}_{g} \gets \texttt{TripContras}(\mathbf{T}_{\texttt{CLS}}, \mathbf{P}_{\texttt{CLS}}, \mathbf{F}_{\texttt{CLS}})$}

\AlgCommentInLine{Subject of interest loss}
\STATE{$\mathcal{L}_{m} \gets 0$}
\FOR{$k'=1, \cdots, k$}
    \STATE{$\mathbf{T}_{s_k}, \mathbf{P}_{s_k}, \mathbf{F}_{s_k}  \gets \mathbf{T}_{\texttt{SOI}}\left[k'\right], \mathbf{P}_{\texttt{SOI}}\left[k'\right],
    \mathbf{F}_{\texttt{SOI}}\left[k'\right]$}

    \STATE{$\mathcal{L}_{m} \gets \mathcal{L}_{m} + G\left[k'\right] * \texttt{WSOILoss}(\mathbf{T}_{s_k}, \mathbf{P}_{s_k}, \mathbf{F}_{s_k})$}

\ENDFOR
\STATE{$\mathcal{L} = \lambda_g \mathcal{L}_g + \lambda_m \mathcal{L}_m$}
\STATE{\textbf{return}  $\mathcal{L}$}
\end{algorithmic}
\end{algorithm}

Let $\mathcal{M}_s$ be a set of subjects within a visual stimulus that the participant is focusing on. It is crucial to recognize that $\mathcal{M}_s$ varies for each individual. For instance, in a stimulus containing both dogs and cats, a person fond of dogs will focus solely on the regions containing dogs, while cat lovers will concentrate on the cats' regions. Consequently, identifying $\mathcal{M}_s$ poses a challenge since we need to possess this predefined set of subjects of interest in advance. To address this challenge, we introduce the SOIP module to identify $\mathcal{M}_s$. 

Since $\mathbf{T}_{\texttt{CLS}}$ encapsulates the entire context of the visual input $x$, we employ this feature as the input of SOIP for predicting which token $t_i \in t$ signifies the interested subject. To achieve this, we establish a trainable matrix $W_m \in \mathbb{R}^{N_c \times d}$ to transform $\textbf{T}_{\texttt{CLS}}$ into a logit vector $z \in \mathbb{R}^{1 \times N_c}$. Each element $z_i \in z$ corresponds to the probability of token $t_i$ representing a subject of focus. Considering the possibility of multiple interested subjects, we utilize the $\texttt{sigmoid}$ function on $z$ instead of $\texttt{softmax}$. Here, $k$ denotes the potential number of interested subjects within the visual input $x$, and we utilize the $\texttt{topK}$ function to select the $k$ tokens with the highest likelihood of representing interested subjects. Mathematically, the SOIP is described as in Eq. \eqref{eq:msp}.
\begin{equation}
\label{eq:msp}
\begin{split}
    z &= \texttt{sigmoid}(\textbf{T}_{\texttt{CLS}} \otimes W_m^{\top}) \quad \text{, } W_m \in \mathbb{R}^{N_c \times d}, z \in  \mathbb{R}^{1\times N_c} \\
    G, I &= \texttt{TopK}(z, k) \quad \text{, } G \in \mathbb{R}^{k}, I \in \mathbb{N}^{+k} \\
    \mathbf{T}_{\texttt{SOI}} &= \left[\textbf{T}_i\right]_{i\in I}
\end{split}
\end{equation}
Where $I$ is a vector stating the location of the token in the $t$ and $G$ is a vector representing the corresponding probability. $\mathbf{T}_{\texttt{SOI}}$ is the a feature set of $\mathcal{M}_s$ where each element is the token feature $\textbf{T}_i$ as defined in the Eq. \eqref{eq:text_encoder} where $i \in I$. The visualization of the SOIP module is illustrated in Fig. \ref{fig:msp}.

\begin{remark}
    \label{rm3}
    The $\mathbf{P}_{s_k}$, $\mathbf{F}_{s_k}$ and $\mathbf{T}_{s_k}$ are features of the same subject. They are equivalent in subject representation if aligned in the same domain space.
\end{remark}

\subsection{Subject of Interest Retrieval}
\label{subsec:soir}

As outlined in the previous section, it is infeasible to directly form $\textbf{P}_{s_k}$ and $\textbf{F}_{s_k}$. The Remark \ref{rm1} and Remark \ref{rm2} imply that if we can find the coefficient $\alpha_{ik}$, we can measure them directly. Fortunately, according to the Remark \ref{rm3}, $\mathbf{P}_{s_k}$, $\mathbf{F}_{s_k}$, and $\mathbf{T}_{s_k}$ are equivalent, but in the form of different modalities. Therefore, we measure the coefficient $\alpha_{ik}$ by measuring the similarity of $\mathbf{T}_{s_k}$ and each token.
Motivated by this idea, we introduce the SOIR module to construct features of interest for subjects in both visual and fMRI modalities. We detail how this module operates for visual $\textbf{P}_{s_k}$ and note that the approach is similarly applicable for $\textbf{F}_{s_k}$.

Given a sequence of visual tokens $\mathbf{P}_{v} = \left[\mathbf{P}_0, \dots, \mathbf{P}_i, \dots, \mathbf{P}_{N_p - 1}\right]$. Firstly, we measure coefficient $\alpha_{ik} = \texttt{sim}(\mathbf{P}_{i}, \textbf{T}_{s_k})$. After that, we use the softmax function over these coefficients to normalize the feature aggregation. Consequently, we form $\textbf{P}_{s_k}$ as the weighted sum of patches' features as mentioned in the Section \ref{subsec:fmri_encoder}. The visualization of the SOIR module is illustrated in Fig. \ref{fig:msp}.

\subsection{Objective Loss Functions}
\label{subsec:objective_loss_functions}
\textbf{Global Contextual Loss}. This loss function maintains a similar global context sharing between visual, text, and fMRI data. In particular, let $\texttt{Contras}$ be a contrastive loss function. The global context loss $\mathcal{L}_{g}$ is defined as in Eq. \eqref{eq:gloabl_contextual_loss}.
\begin{equation}
\label{eq:gloabl_contextual_loss}
\begin{split}
    \mathcal{L}_{g} &= \texttt{Contras}(\textbf{T}_{\texttt{CLS}}, \textbf{P}_{\texttt{CLS}}) \\
    &+ \texttt{Contras}(\textbf{T}_{\texttt{CLS}}, \textbf{F}_{\texttt{CLS}}) \\
    &+ \texttt{Contras}(\textbf{F}_{\texttt{CLS}}, \textbf{P}_{\texttt{CLS}}) 
    \\
    \texttt{Contras}(x, y) &= -\frac{1}{N}\sum_i^N \log\frac{\exp({x}_i \otimes {y}_i/\sigma)}{\sum_j^N\exp({x}_i \otimes {y}_j/\sigma)} \\
    &- \frac{1}{N}\sum_i^N \log\frac{\exp({y}_i \otimes {x}_i/\sigma)}{\sum_j^N\exp({y}_i \otimes {x}_j/\sigma)}
\end{split}
\end{equation} 

\textbf{Weighted SOI Loss (WSOILoss)}.
With the SOIP and SOIR modules, we get the features of the interested subjects in text, visual, and fMRI domains, respectively. It is crucial to learn similarities among them from the same subject and to distinguish them from those of distinct subjects. Similar to the global contextual loss mentioned earlier, we employ a contrastive loss function for each proposed subject. However, unlike the equation mentioned earlier, we incorporate a loss weight for each proposal. It is because, with a fixed number $k$ of proposed subjects, there may be instances where the number of minding subjects is less than $k$. Relying solely on the confidence score as a weight for the loss should aid SOIP in better learning and in reducing false positives in that module. The Weighted SOI Loss $\mathcal{L}_m$ is formulated as in Eq. \eqref{eq:minding_subject_loss}.
\begin{equation}
\label{eq:minding_subject_loss}
\begin{split}
    \mathcal{L}_m = \sum_{i=0}^{k-1}G\left[i\right] \Big[ &\texttt{Contras}(\mathbf{T}_{s_i}, \mathbf{P}_{s_i}) + \\ &\texttt{Contras}(\mathbf{T}_{s_i}, \mathbf{F}_{s_i}) +  \\ &\texttt{Contras}(\mathbf{F}_{s_i}, \mathbf{P}_{s_i})\Big]
\end{split}
\end{equation}
In summary, the BRACTIVE is trained by the following loss function: 
$\mathcal{L} = \lambda_g \mathcal{L}_g + \lambda_m \mathcal{L}_m$, 
where $\lambda_g$ and $\lambda_m$ are the weights for two corresponding loss functions. Overall, the proposed method is described in the Algorithm \ref{algo:BRACTIVE}.

\section{Brain Region of Interest Localization}
\label{sec:detecting_fmri_signal}
\begin{algorithm}[!h]
\centering
\small
\caption{Pseudo code for ROIs detection.}
\label{algo:detecting_fmri_signal}
\begin{algorithmic}[1]
\STATE {\bfseries Input:} The features of the interested subject $\mathbf{F}_{s_k} \in \mathbb{R}^{1\times d}$, patches' features of flattened fMRI 2D $\mathbf{F}_{f}\in \mathbb{R}^{N_r \times d}$, \\
scale factor $s$, threshold $\gamma$
\STATE {\bfseries Output:} The optimization loss function.

\STATE{$\mathbf{F}_{s_k} \gets \texttt{l2norm}(\mathbf{F}_{s_k})$}
\STATE{$\mathbf{F}_{f} \gets \texttt{l2norm}(\mathbf{F}_{f})$}
\STATE{$B \gets \mathbf{F}_{s_k} \otimes \mathbf{F}_{f}^{\top} \in \mathbb{R}^{1 \times N_r}$}
\STATE{$B \gets \texttt{upsampling}(B, s) \in \mathbb{R}^{h_{2d} \times w_{2d}}$}
\STATE{$B \gets \texttt{T}_{2d}^{\top}(B) \in \mathbb{R}^{N_f}$}
\STATE{$B \gets B > \gamma$}
\STATE{\textbf{return}  $B$}
\end{algorithmic}
\end{algorithm}

As noted in Remark \ref{rm2}, the sub-signals or voxels in fMRI corresponding to a subject can be identified by evaluating the coefficient $\beta_{ik}$. This coefficient is defined as $\beta_{ik} = \texttt{sim}(\mathbf{F}_{s_k}, \mathbf{F}i)$, representing the cosine similarity between $\mathbf{F}_{s_k}$ and $\mathbf{F}_i$. Additionally, a threshold $\gamma$ is predefined, designating that a sub-signal $f_i \in f$ is associated with the subject $s_k$ if $\beta_{ik} > \gamma$. The procedure for detecting the subject-specific fMRI signal is detailed in Algorithm \ref{algo:detecting_fmri_signal}. This algorithm highlights two principal observations. (1) Thanks to the Weighted SOI Loss, we can employ both kinds of features from the interested subjects, such as $\mathbf{T}_{s_k}, \mathbf{P}_{s_k}$, as inputs of the algorithm. This flexibility enables the use of multiple approaches in studying brain activities. Either text or vision modalities can be used to detect the ROIs of the subjects. (2) The Algorithm \ref{algo:detecting_fmri_signal} can be adapted for visual analysis by substituting $\textbf{F}_f$ with $\mathbf{P}_v$. This modification allows us to identify which parts of an image strongly correlate with predetermined regions of interest (ROIs) in the brain.

\section{Implementation Details}
\label{sec:implementation}
\textbf{Visual Encoder and fMRI Encoder}. We employ the ViT-B-16 for both the visual and fMRI encoders. The input image size is $224 \times 224$. The feature dimension of these encoders is set to 1024.
\\ \noindent 
\textbf{Text Encoder}. We utilize the pre-trained transformer architecture from \cite{clip} to extract features from the text. The maximum length of the context is set to 77 \cite{clip} and the dimension of text features is also set to 1024. The text encoder is frozen while training. It will help to align visual and fMRI features with the text features easily. 
\\ \noindent 
\textbf{SOIP}. For the SOIP module, we select $k=4$ while proposing the potential subject of interest.
\\ \noindent
\textbf{Datasets.} To train BRACTIVE, we use the Natural Scenes Dataset (NSD) \cite{allen2022massive}, a comprehensive compilation of responses from eight participants obtained through high-quality 7T fMRI scans. Each subject was exposed to approximately 73,000 natural scenes, forming the basis for constructing visual brain encoding models. Since the visual stimulus in this database is a subset of COCO \cite{coco}, for each sample, there exist captions (text modality), the visual stimulus, and a corresponding fMRI response. For each subject, we split the data into five folds for training and validation.
\\ \noindent
\textbf{Training Strategy}. BRACTIVE is implemented in the PyTorch framework and trained using 16 $\times$ A100 GPUs (40G each). The learning rate is initially set to $2.5e^{-5}$ and then gradually reduced to zero under the CosineLinear policy \cite{cosine}. The batch size is set to $4$/GPU. The model is optimized by AdamW \cite{adamw} for $30$ epochs. The training is completed within two hours.

\section{Experimental Results}
\label{sec:results}

\rebuttaltpami{In this section, we firstly evaluate the accuracy of the proposed BRACTIVE in localizing ROIs w.r.t human subject in Section \ref{subsec:exp_localized_roi}. Next, we evaluate the effectiveness of the method on downstream tasks in Section \ref{subsec:brain_response_prediction} and Section \ref{subsec:other_downstream}.
}

\subsection{Localized ROIs w.r.t Human}
\label{subsec:exp_localized_roi}
\begin{table*}[!h]
    \centering
    \caption{The Dice Score of ROIs localization w.r.t human subjects. The \textit{AvgAtt} is the average of the attention maps of human subjects across participants. \rebuttaltpami{The asterisk denotes our method trained with captions generated by the Vision-Language Model. The \textbf{bold} indicates the best results.}}
    \vspace{-4mm}
    \resizebox{1.0\textwidth}{!}{
    \begin{tabular}{l|c|c|c|c|c|c|c|c|c}
        \hline
        Method &  P1 &  P2&  P3&  P4&  P5& P6 & P7 & P8 & AvgAtt \\
        \hline
        GradCAM \cite{grad_cam} &  57.36($\pm$ 0.3)&  56.73($\pm$ 0.1)&  58.32($\pm$ 0.2)&  55.41($\pm$ 0.2)&  60.78($\pm$ 0.1)&  61.24($\pm$ 0.1)& 59.29($\pm$ 0.3) & 56.52($\pm$ 0.2) & 54.84($\pm$ 0.2) \\
        Ours &  {68.98($\pm$ 0.2)}&  {69.32($\pm$ 0.2)}&  {69.31($\pm$ 0.3)}&  {69.84($\pm$ 0.1)}&  {69.31($\pm$ 0.3)}&  {69.70($\pm$ 0.1)}& \textbf{69.87($\pm$ 0.1)} & {69.37($\pm$ 0.3)} & {72.57($\pm$ 0.2)}\\
        \rebuttaltpami{Ours$^*$} &  \rebuttaltpami{\textbf{69.10($\pm$ 0.3)}}&  \rebuttaltpami{\textbf{69.47($\pm$ 0.2)}}&  \rebuttaltpami{\textbf{69.50($\pm$ 0.1)}}&  \rebuttaltpami{\textbf{69.92($\pm$ 0.2)}}&  \rebuttaltpami{\textbf{69.56($\pm$ 0.2)}}&  \rebuttaltpami{\textbf{69.80($\pm$ 0.1)}}&  \rebuttaltpami{{69.82($\pm$ 0.3)}}&  \rebuttaltpami{\textbf{69.38($\pm$ 0.2)}}&  \rebuttaltpami{\textbf{72.80($\pm$ 0.2)}}\\
        \hline
    \end{tabular}
    }
    \label{tab:fmri_detection}
\end{table*}

\begin{table*}[!ht]
    \centering
    \caption{Performance of the brain response prediction. IMN1K refers to pretrained weights from Imagnet1K \cite{imagenet}. CLS and SOI refer to the CLS token and Subject of Interest features, respectively. \rebuttaltpami{The asterisk denotes our method trained with captions generated by the Vision-Language Model. The \textbf{bold} indicates the best results.}}
    \vspace{-4mm}
    \resizebox{1.0\textwidth}{!}{
    \begin{tabular}{l|c|c|c|c|c|c|c|c}
        \hline
        Pretrained &  P1 &  P2&  P3&  P4&  P5& P6 & P7 & P8\\
        \hline
        IMN1K \cite{imagenet} & 41.69($\pm$ 0.3) & 41.95($\pm$ 0.3)& 41.52($\pm$ 0.4)& 41.81($\pm$ 0.4)& 41.35($\pm$ 0.2)& 41.17($\pm$ 0.3)& 41.40($\pm$ 0.3)& 41.02($\pm$ 0.2) \\
        Ours + CLS &  47.43($\pm$ 0.3)& 46.94($\pm$ 0.3)& 47.28($\pm$ 0.2)& 47.91($\pm$ 0.3)& 46.94($\pm$ 0.4)& 46.66($\pm$ 0.4)& 47.31($\pm$ 0.3)& 47.22($\pm$ 0.3) \\
        Ours + SOI &  {50.31($\pm$ 0.4)}& {50.70($\pm$ 0.2)} & {51.19($\pm$ 0.3)}& {51.32($\pm$ 0.3)}& {50.79($\pm$ 0.3)}& {50.64($\pm$ 0.3)}&{51.25($\pm$ 0.4)}& {51.04($\pm$ 0.4)} \\
        \rebuttaltpami{Ours$^*$ + SOI} & \rebuttaltpami{\textbf{50.52($\pm$ 0.3)}}&  \rebuttaltpami{\textbf{50.84($\pm$ 0.2)}}&  \rebuttaltpami{\textbf{51.48($\pm$ 0.2)}}&  \rebuttaltpami{\textbf{51.46($\pm$ 0.3)}}&  \rebuttaltpami{\textbf{50.88($\pm$ 0.2)}}&  \rebuttaltpami{\textbf{50.93($\pm$ 0.2)}}&  \rebuttaltpami{\textbf{51.72($\pm$ 0.3)}}&  \rebuttaltpami{\textbf{51.19($\pm$ 0.3)}} \\
        \hline
    \end{tabular}
    }
    \label{tab:brain_response_prediction}
\end{table*}

\begin{figure*}[!ht]
    \centering
    \includegraphics[width=1\linewidth]{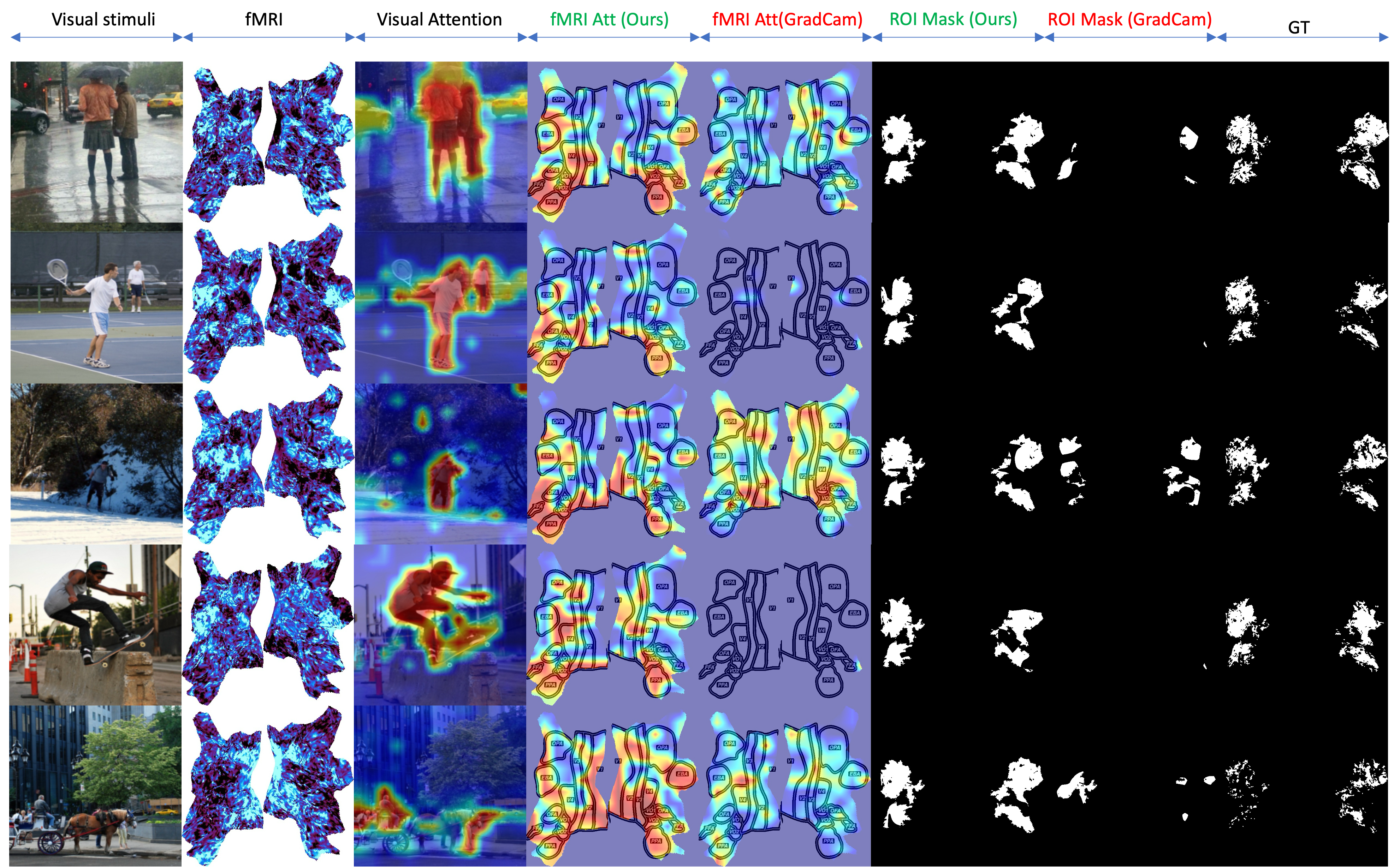}
    \vspace{-4mm}
    \caption{ROI Mask w.r.t Human. The mask is generated by applying 0.5 of the threshold to the attention maps. The $6^{th}$ and $7^{th}$ columns are the ROI masks of our method and GradCam, respectively. The last column is the ground truth of ROIs w.r.t human. }
    \label{fig:ab_human_rois}
    \vspace{-4mm}
\end{figure*}

Since the ROIs for specific subjects are limited, in the scope of this paper, we evaluate the accuracy of BRACTIVE with respect to human subjects only. In particular, we obtain the ROIs of the \textit{floc-faces} and \textit{floc-bodies} areas inside the brain as the ground truth and evaluate how well our algorithm \ref{algo:detecting_fmri_signal} can localize the ROIs of the human subjects. \rebuttaltpami{In the NSD dataset, ground-truth ROIs are obtained by having each participant perform category functional localizer (fLoc) experiments \cite{stigliani2015temporal}. These brain regions' annotations and activations can vary across individuals due to anatomical and functional differences.}.  We utilize dice scores as the evaluation metric. The results are demonstrated in the Table \ref{tab:fmri_detection}. The details and visualization of the prediction and ground truth can be found in the \rebuttaltpami{Section \ref{sec:qualitative_human_roi} and Fig. \ref{fig:ab_human_rois}}.

\begin{figure*}[!ht]
    \centering
    \includegraphics[width=0.95\linewidth]{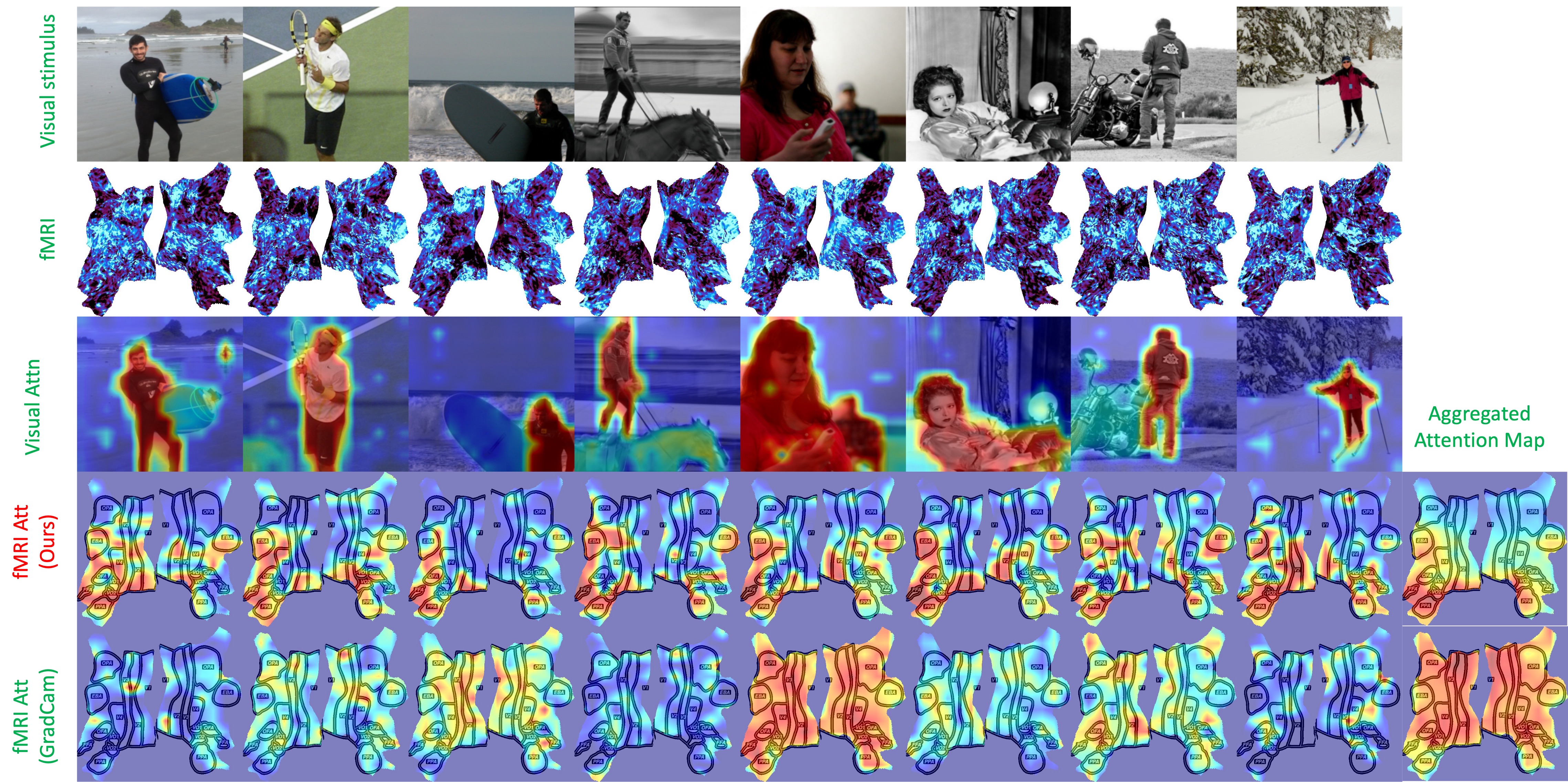}
    \vspace{-4mm}
    \caption{Region Of Interest of the human subject perceived by participant P1. The last column includes the mean of attention maps across all samples of all participants in the NSD. In the attention map, the brighter the colors, the higher the correlation to the subject.}
    \label{fig:fmri_detection_human}
\end{figure*}

To establish a baseline for ROI localization, we employ ViT-B-16 as an fMRI encoder. We develop a classification model that processes fMRI signals and predicts the subject categories depicted in the visual stimuli. Since the visual stimulus from NSD belongs to the COCO database, our model targets classification into 80 fixed categories. Upon completing the training, we utilize GradCAM \cite{grad_cam} to identify which regions correspond to specific subjects, such as humans or persons. We set a threshold in GradCAM to define the semantics of the ROIs. This baseline archives the dice score from 55.41\% to 61.24\% for eight participants. 

In the second experiment, we use the steps outlined in Algorithm \ref{algo:detecting_fmri_signal} to localize the ROIs. Overall, we observe a notable enhancement ranging from 68\% to 69\% consistently across eight participants, compared to GradCam \cite{grad_cam}. In the third experiment, we take the average of the attention maps across all participants. Surprisingly, the performance drops to 54.84 \% for the GradCam \cite{grad_cam} baseline while we observe a notable improvement of 72.57\% from our method.

\begin{table}[!hb]
\centering
\caption{Results of the downstream tasks: object detection, instance segmentation on the COCO dataset, and semantic segmentation on the ADE20K dataset}
\label{tab:coco_object_detection}
\vspace{-4mm}
\begin{tabular}{c|c|c|c}
\Xhline{1.0pt}
 Pretrain & AP$^\text{box}$ & AP$^\text{segm}$ & mIoU \\
\hline
 IMN1K \cite{imagenet} & {49.8($\pm$ 0.2)} & {44.5($\pm$ 0.3)} & {46.4($\pm$ 0.3)} \\
CLIP \cite{clip} & {50.2($\pm$ 0.4)} & {44.9($\pm$ 0.3)} & {47.6($\pm$ 0.4)} \\
Ours & \textbf{51.4($\pm$ 0.3)} & \textbf{45.7($\pm$ 0.2)} & \textbf{48.7($\pm$ 0.2)} \\
 \hline
\end{tabular}
\end{table}

\subsection{Brain Response Prediction}
\label{subsec:brain_response_prediction}

This downstream task aims to predict the neural responses of the human brain to natural scenes, as observed during participant viewing sessions \cite{algonauts, nsd}. We employ the pretrained visual encoder $\mathcal{E}_v$ to fine-tune this task. Our evaluation follows the protocol described in \cite{algonauts}, using the Pearson Correlation Coefficient (PCC) as the metric. Results, as presented in Table \ref{tab:brain_response_prediction}, compare different architectures and training strategies. 
The performance for these methods on eight participants varies between 41.02 and 41.95. In a second experiment, we use the pretrained visual encoder from BRACTIVE as initial weights and employ the features from the \texttt{CLS} token in the final linear layer, thereby enhancing performance by approximately 5\% compared to the initial scenario. In a final experiment, settings remain similar to the second, but instead of using \texttt{CLS} token features, we use a weighted sum of SOI features. This adjustment leads to substantial performance improvements of 10\% and 3\% over the first and second experiments, respectively. The SOI features provide a richer context within the visual stimuli than the \texttt{CLS} features, explaining the observed enhancements in model performance.

\subsection{Other Downstream Tasks: Detection and Segmentation}
\label{subsec:other_downstream}

This section examines how the pre-trained vision encoder $\mathcal{E}_v$ can assist with other downstream tasks. Specifically, we select the most common vision tasks: object detection, instance segmentation, and semantic segmentation. For object detection and instance segmentation, we utilize the Mask-RCNN \cite{maskrcnn} architecture and the COCO \cite{coco} database for training and evaluation. For semantic segmentation, we utilize the UpperNet \cite{uppernet} architecture and the ADE20K \cite{ade20k} database. We conveniently utilize the base code from MMDetection and MMSegmentation \cite{mmdetection}. The performance is shown in the Table \ref{tab:coco_object_detection}.

The key finding is that: \textit{Human brain activities significantly contribute a step toward human-like capabilities}. Indeed, taking pretrained vision encoder $\mathcal{E}_v$ from BRACTIVE as the backbone, we observe improvements by all tasks compared to the one that pretraining by CLIP, IMN1K, and  51.4\% of AP$^\text{box}$, 45.7\%of AP$^\text{segm}$, and 48.7\% of mIoU, respectively. These results are higher than those of other pre-trained methods \cite{clip, imagenet}.

\begin{figure*}[!ht]
    \includegraphics[width=1.0\linewidth]{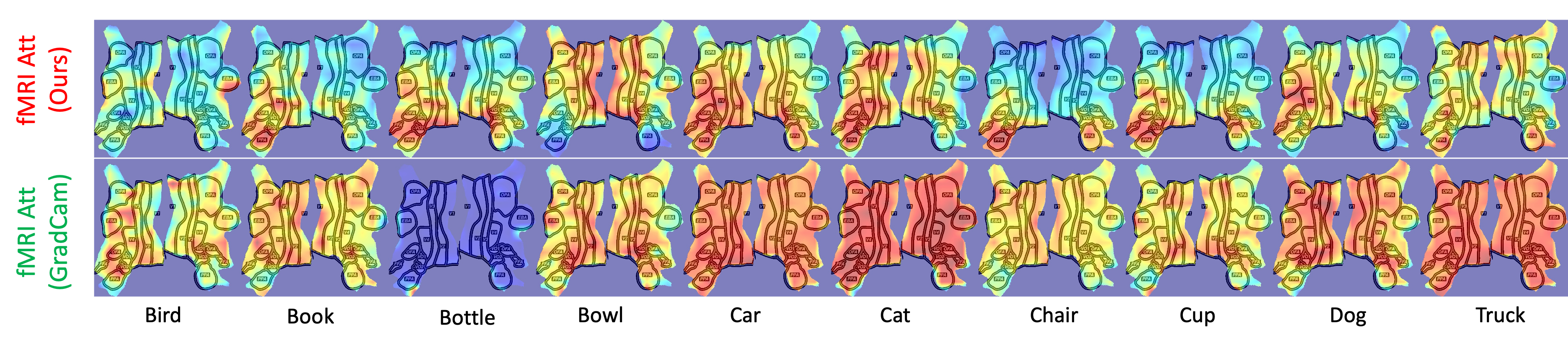}
    \vspace{-10mm}
    \caption{Average ROIs of the non-human subjects perceived by participant P1.}
    \label{fig:fmri_detection_non_human}
\end{figure*}

\begin{figure*}[!ht]
    \centering
    \includegraphics[width=1\linewidth]{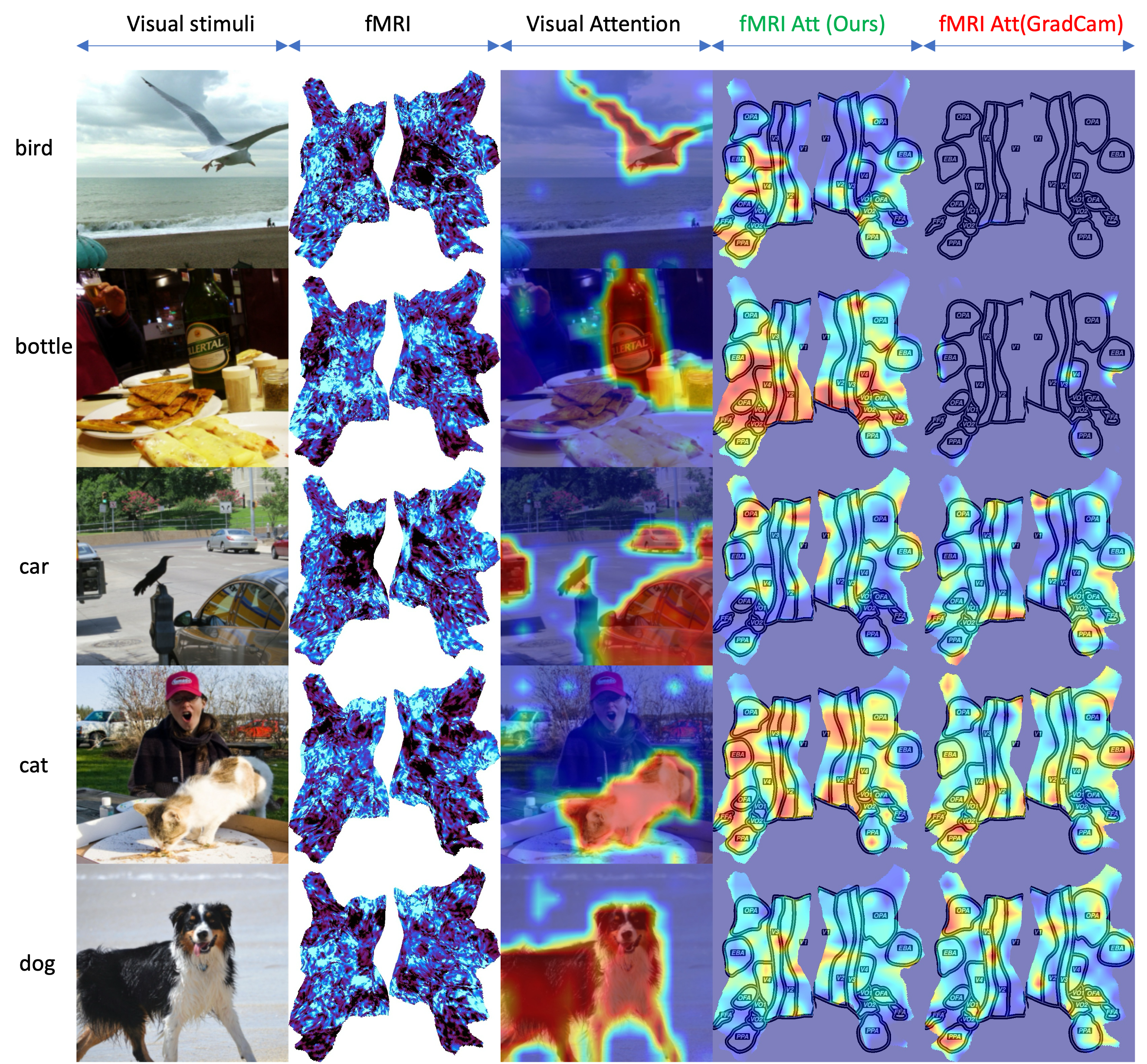}
    \caption{ROI w.r.t Non-human Subjects.}
    \label{fig:ab_non_human_rois}
\end{figure*}

\section{Ablation Studies}
In this section, we demonstrate the detected ROIs of both human and non-human subjects and then analyze the meaning behind them. \rebuttaltpami{In addition, we also analyze the performance of the proposed method trained on the dense captions generated by Vision-Language Model (VLM).}

\subsection{Qualitative Results of Human Subjects' ROI}
\label{sec:qualitative_human_roi}

We illustrate the brain region attention of the human subject as in Fig. \ref{fig:fmri_detection_human}. The first row displays the visual stimulus, while the second row illustrates the corresponding brain activities, a 2D version of the fMRI signal. The third row shows the visual attention map, which contains similarity scores between the  SOI features and each image patch. Likewise, the fourth row displays the attention map for the SOI features in relation to each patch of brain activities. The final row is the attention map generated by GradCam \cite{grad_cam}. 

Our analysis yielded two principal observations. Firstly, the attention map of the visual SOI features prominently focuses on the human figure within the visual stimulus. The visual attention map shares the same role as the fMRI attention map because they are generated by the same algorithm. Therefore, the fMRI attention map also potentially shows the right ROI. Secondly, detected ROIs show a consistent form across different visual stimuli. In particular, most attention scores are highly activated in the OFA, PPA, and FFA regions, which respond to human face and body processing.

\rebuttaltpami{
In addition, we further illustrate the ROI mask of the human subject as in Fig. \ref{fig:ab_human_rois}. The ROI masks are constructed by applying a threshold, i.e., 0.5, to the attention map. The last column is the ground truth mask, which shows the brain activations of the floc-bodies and floc-faces ROIs. The $6^{th}$ column is the ROI mask of our proposed method, while the $7^{th}$ column is the ROI mask of GradCam \cite{grad_cam}. We measure the dice score between the predicted masks and the corresponding ground truth. The performance is reported in the Table \ref{tab:fmri_detection}. The figure demonstrates that the ROI masks generated by our approach are much closer to the ground truth than the ROI masks produced by GradCam \cite{grad_cam}, which are mostly empty.
}

\subsection{Qualitative Results of Non-Human Subjects' ROI}

We illustrate the average attention maps generated by our method and GradCam for non-human subjects, as shown in Fig. \ref{fig:fmri_detection_non_human}. The key finding is that: our approach generates different attention maps for various subjects, providing valuable insights for neuroscientists. Indeed, the GradCam \cite{grad_cam} generates similar attention maps for car, cat, dog, and truck subjects, while it also could not determine ROIs for bottles. 

\rebuttaltpami{
We further illustrate the ROIs of non-human subjects in Fig. \ref{fig:ab_non_human_rois}. Two key findings emerge from our observations. First, the visual attention maps generated by our method are well-fitted to the subjects. Second, compared to GradCam, our method localizes the ROIs more efficiently. For instance, with bird and bottle subjects within the visual stimulus, GradCam appears to have failed in detecting the ROIs, whereas our method succeeds.
}

\subsection{Caption Dependence of the SOIP Module}
\label{sec:ablation_caption}
\color{black}
One potential limitation of BRACTIVE is its reliance on text captions to infer the subjects of interest through the SOIP module. The captions provided in the NSD dataset (adapted from MS-COCO) may omit objects that are present in the visual stimuli or contain ambiguous descriptions due to participant distraction. To investigate the impact of caption quality on BRACTIVE, we conducted an ablation study comparing different sources of captions.
\\
\noindent
We replaced the original NSD-provided COCO captions with automatically generated captions using BLIP-2 \cite{li2023blip}, a state-of-the-art Vision-Language Model capable of producing dense image descriptions. Unlike COCO captions, which typically mention one or two salient objects, BLIP-2 \cite{li2023blip} captions cover a broader set of fine-grained objects and contextual details within the visual scene. BRACTIVE was trained with both types of captions under the same experimental protocol, and performance was measured on the standard ROI prediction tasks.
\\
\noindent
As shown in Table \ref{tab:fmri_detection} and Table \ref{tab:brain_response_prediction}, BRACTIVE trained with BLIP-2 \cite{li2023blip} captions achieves competitive performance compared to the version trained with COCO captions, with marginal differences across metrics (within 0.3\%). This indicates that dense captions generated by VLMs are a viable substitute when ground-truth captions are incomplete or unavailable. Moreover, the ability to generate captions for any new visual stimuli relaxes the dependence on manually annotated captions and broadens the applicability of BRACTIVE to datasets beyond NSD.
\color{black}

\section{Conclusion, Limitations and Broader Impacts}
\label{sec:conclusion}

This paper introduces the new BRACTIVE framework to automatically localize the visual brain region of interest. The experimental results demonstrate that the ROIs with respect to human subjects identified by BRACTIVE are highly correlated with the predefined ROIs from neuroscience. In addition, BRACTIVE demonstrates its superiority by identifying ROIs in various non-human subjects. This ability provides reasonable references for neuroscience fields, facilitating further discovery of the human brain. Finally, BRACTIVE promises a step forward in human-like learning by leveraging brain activities to guide the DNNs, requiring less effort in data collection and yielding higher performance gains. 

\noindent \textbf{Limitations.} 
\rebuttaltpami{
In our proposed method, text is leveraged to provide contextual descriptions of the visual stimulus. However, these captions may not always include the subject that the participant is focusing on, or may misalign with the brain responses in cases where the participant was distracted during the experiment. To examine this limitation, we conducted an ablation study (Section~\ref{sec:ablation_caption}) where captions were replaced by dense descriptions automatically generated with a vision-language model. The results showed that such automatically generated captions can serve as a competitive alternative, thereby partially mitigating the reliance on dataset-provided text. This simple solution highlights the potential of reducing caption dependence through automated or multimodal captioning strategies. Nevertheless, a more ambitious direction is to reduce reliance on text altogether by developing methods that rely purely on visual and neural signals. We leave this as an interesting direction for future work.
}

\noindent \textbf{Future Work \& Broader Impacts.} We plan to address the limitations above and extend the BRACTIVE to different brain functionalities such as language processing, emotions, \rebuttaltpami{abstract art, dynamic video stimuli, or larger participant cohorts}. More importantly, BRACTIVE promises to be a reference tool for neuroscientists, supporting them to \textit{debug} brain functions. \rebuttaltpami{We see this as a promising direction for future collaborations with cognitive neuroscience labs}. In addition, with a deeper understanding of how the brain works, further novel DNNs will be inspired by these findings, boosting the AI field.

\section{Acknowledgment}
\rebuttaltpami{
This research is supported by the Arkansas High Performance Computing Center, which is funded through multiple National Science Foundation grants and the Arkansas Economic Development Commission.}

\bibliographystyle{IEEEtran}
\bibliography{egbib}

\vspace{-1.5cm}
\begin{IEEEbiography}[{\includegraphics[width=1in,height=1.25in,clip,keepaspectratio]{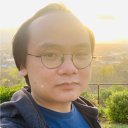}}]{Xuan-Bac Nguyen}
is currently a Ph.D. student at the Department of Computer Science and Computer Engineering of the University of Arkansas. He received his M.Sc. degree in Computer Science from the Electrical and Computer Engineering Department at Chonnam National University, South Korea, in 2020.  He received his B.Sc. degree in Electronics and Telecommunications from the University of Engineering and Technology, VNU, in 2015. In 2016, he was a software engineer in Yokohama, Japan. His research interests include Quantum Machine Learning, Face Recognition, Facial Expression, and Medical Image Processing.
\end{IEEEbiography}
\vspace{-1.0cm}

\begin{IEEEbiography}
[{\includegraphics[width=1in,height=1.25in,clip,keepaspectratio]{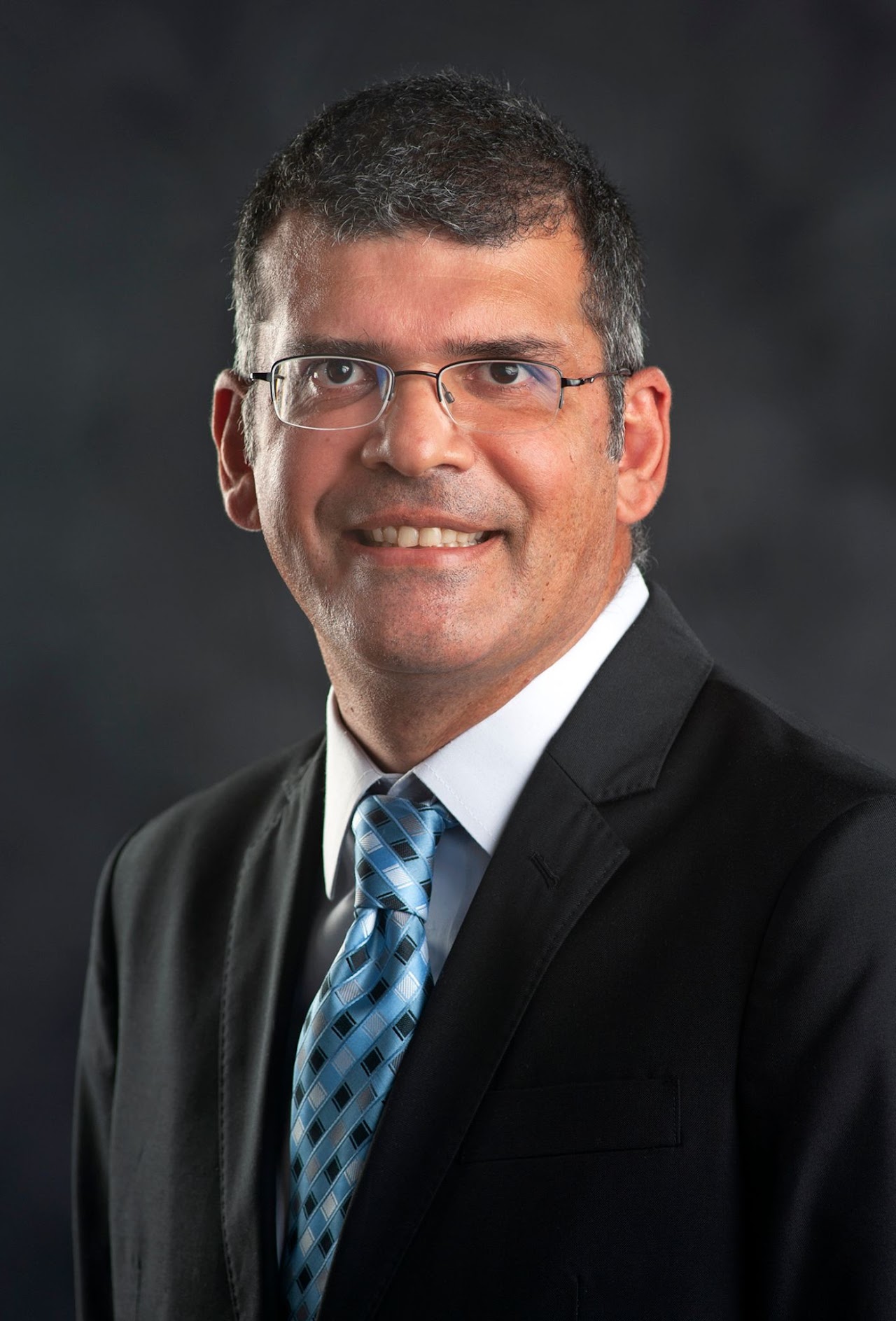}}]{Samee U. Khan} received a Ph.D. in 2007 from the University of Texas, Arlington, TX. He is the George J. and Alice D. Fiedler Distinguished Chair Professor \& Head in the Mike Wieges Department of Electrical and Computer Engineering at Kansas State University (K-State). Before joining K-State, he was a faculty member at Mississippi State University (MSU), serving as Department Head and the James W. Bagley Chair Professor from 2020 to 2024. He started his career at North Dakota State University in 2008 and rose through the ranks to become the Walter B. Booth Professor. While at NDSU, he was assigned to the National Science Foundation (2016-2020) as Cluster Lead for Computer Systems Research within the Computer and Network Systems Division. 
His research interests include computer system optimization, robustness, and security. His work has appeared in over 475 publications. He is the associate editor of IEEE Transactions on Cloud Computing and the Journal of Parallel and Distributed Computing. He is a Fellow of the IET and BCS, a Distinguished Member of the ACM, and a Senior Member of the IEEE. 
He has won several awards, including the Best Paper Award (Systems Track), IEEE Cloud Summit, 2024; IEEE R3 Outstanding Engineer Award, 2024; IEEE Computer Society Distinguished Contributor Award, 2022 (inducted in the inaugural class); IEEE ComSoc Technical Committee on Big Data Best Journal Paper Award, 2019; IEEE-USA Professional Achievement Award, 2016; IEEE Golden Core Member Award, 2016; IEEE TCSC Award for Excellence in Scalable Computing Research (Middle Career Researcher), 2016; IEEE Computer Society Meritorious Service Certificate, 2016; Tapestry of Diverse Talents Award, North Dakota State University (NDSU), ND, USA, 2016; Exemplary Editor, IEEE Communications Surveys and Tutorials, IEEE Communications Society, 2014; Outstanding Summer Undergraduate Research Faculty Mentor Award, NDSU, ND, USA, 2013; Best Paper Award, IEEE Intl. Conf. on Scalable Computing and Communications (ScalCom), 2012; Sudhir Mehta Memorial International Faculty Award, NDSU, ND, USA, 2012; Best Paper Award, ACM/IEEE Intl. Conf. on Green Computing \& Communications (GreenCom), 2010.
\end{IEEEbiography}

\begin{IEEEbiography}[{\includegraphics[width=1in,height=1.25in,clip,keepaspectratio]{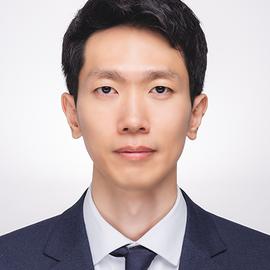}}]{Hojin Jang} is an Assistant Professor in the Department of Brain and Cognitive Engineering at Korea University and the Director of the Human-Machine Cognition Lab. He is also a Research Affiliate in the Department of Brain and Cognitive Sciences at MIT. His research interests encompass Vision Science, Computational Cognitive Neuroscience, Human-AI Cognition, and Brain Decoding. He has been awarded a U.S. patent and the Jum Nunnally Dissertation Award. Dr. Jang is an active member of the Vision Sciences Society and has presented three oral presentations at its conferences.
\end{IEEEbiography}
\vspace{-1.5cm}

\begin{IEEEbiography}[{\includegraphics[width=1in,height=1.25in,clip,keepaspectratio]{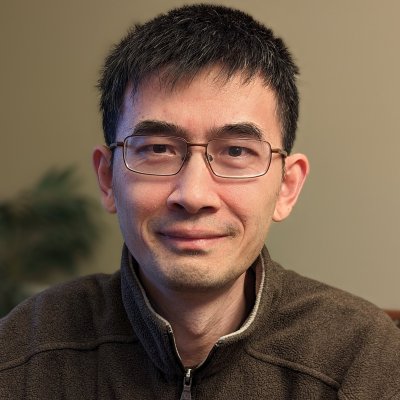}}]{Xin Li} (Fellow, IEEE) received the BS degree with
highest honors in electronic engineering and information science from University of Science and Technology of China, Hefei, in 1996, and the PhD degree in electrical engineering from Princeton University,
Princeton, NJ, in 2000. He was a member of technical staff with Sharp Laboratories of America, Camas, WA from 2000 to 2002. He was a faculty member in Lane Department of Computer Science and Electrical
Engineering, West Virginia University from 2003 to 2023. Currently, he is with the Department of Computer Science, University at Albany, Albany, NY 12222 USA. His research interests include image and video processing, compute vision and computational neuroscience. He was elected a Fellow of IEEE, in 2017 for his contributions to image interpolation, restoration and compression.
\end{IEEEbiography}
\vspace{-1.5cm}

\begin{IEEEbiography}
[{\includegraphics[width=1in,height=1.25in,clip,keepaspectratio]{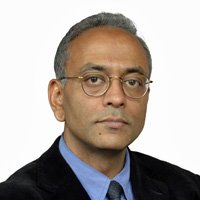}}]{Pawan Sinha} is a professor of neuroscience at MIT. Pawan’s research interests span brain science, AI, and public health. His experimental work involves studying healthy individuals and also those with neurological disorders such as autism. Pawan founded Project Prakash in 2005 with the twin objectives of providing treatment to children with severe visual impairments and also understanding mechanisms of learning and plasticity in the brain. This project has provided insights into several fundamental questions about brain function (even some that had remained open for the past three centuries) while also transforming the lives of many blind children by bringing them the gift of sight.
Pawan is a recipient of the PECASE – US Government’s highest award for young scientists, the Sloan Foundation Fellowship, and the Troland Award from the National Academies. He was inducted into the Guinness Book of World Records for creating the world’s smallest reproduction of a printed book.
\end{IEEEbiography}
\begin{IEEEbiography}
[{\includegraphics[width=1in,height=1.25in,clip,keepaspectratio]{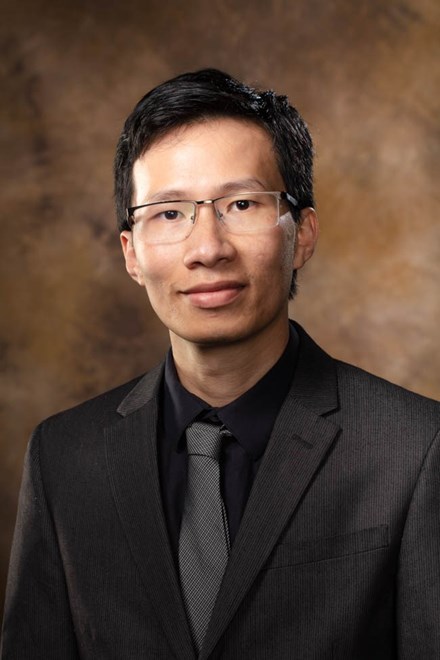}}]{Khoa Luu} is an Associate Professor and the Director of the Computer Vision and Image Understanding (CVIU) Lab in the Department of Computer Science \& Computer Engineering at the University of Arkansas, Fayetteville. He is affiliated with the Center for Public Health and Technology, UA, and the NSF MonARK Quantum Foundry. He is an Associate Editor of the Multimedia Tools and Applications Journal, Springer Nature. He is also the Area Chair in CVPR 2023-2025, NeurIPS 2024-25, WACV 2025, ICML 2025, ICLR 2025-2026, AAAI 2026 and Quantum Week 2025. He was the Research Project Director in the Cylab Biometrics Center at Carnegie Mellon University (CMU), USA. 
He received eight patents, three Best Paper awards, and co-authored over 120 papers in conferences, technical reports, and journals.
He was a vice chair of the Montreal Chapter IEEE SMCS in Canada from September 2009 to March 2011. His research interests include Computer Vision, Semantic Video Understanding, Biometrics, Face Recognition, Tracking, Human Behavior Understanding, Domain Adaptation, Deep Generative Modeling, Image and Video Processing, Compressed Sensing, and Quantum Machine Learning. 
He is a co-organizer and chair of the CVPR Precognition Workshop 2019-2025; the MICCAI Workshop in 2019, 2020, and the ICCV Workshop in 2021. He is a PC member of AAAI, ICPRAI in 2020 and 2022. 

\end{IEEEbiography}

\end{document}